# Recurrent Neural Network on PICTURE Model

By

Weihan Xu

A senior thesis submitted in partial fulfillment of the requirements
for the degree of
Bachelor of Science
(Honor Computer Science)
In the University of Michigan
2023

Advisors:
Dr. Kayvan Najarian, Professor of Computer Science
Dr. Sardar Ansari, Research Assistant Professor of Emergency Medicine

Second Reader:
Dr. David Hanauer, Associate Professor of Informatics

# Table of contents

# Acknowledgements

First and foremost, I would like to express my sincere gratitude to Dr. Kayvan Najarian and Dr. Sardar Ansari for their invaluable guidance and support throughout the process of advising my senior thesis. I would also like to extend my appreciation to Dr. David Hanauer for serving as the second reader for my thesis.

Furthermore, I would like to acknowledge the contribution of Brandon Cummings, who provided me with the general idea of the research project, which served as a foundation for my work. Last but not least, I would like to thank Loc Cao for providing me with valuable insights into research directions and helping me with the implementation of the code.

# Abstract

Background:
    Intensive Care Units (ICUs) provide critical care and life support for most severely ill and injured patients in the hospital. With the need for ICUs growing rapidly and unprecedentedly, especially during COVID-19, accurately identifying the most critical patients helps hospitals to allocate resources more efficiently and save more lives. The Predicting Intensive Care Transfers and Other Unforeseen Events (PICTURE) model predicts patient deterioration by separating those at high risk for imminent intensive care unit transfer, respiratory failure, or death from those at lower risk.

Objective:
    This study aims to implement a deep learning model to benchmark the performance from the XGBoost model, an existing model which has competitive results on prediction.

Data:
    The windowed dataset comprises 132,589 patients with a total of 223,824 encounters. The dataset also includes a total of 3,349,770 observations. Similarly, the granular dataset consists of 132,474 patients with a total of 223,640 encounters. In total, the granular dataset includes 17,452,991 observations.

Methods:
    The PICTURE dataset was used to train and validate the model. Three different approaches were employed to preprocess the data: 1) a sliding window, 2) a dense sliding window, and 3) smart batching method. Various deep learning models, such as Recurrent Neural Network (RNN) and transformers, were then trained on the preprocessed data.
    Finally, the performance of all models was evaluated based on accuracy and threshold-independent metrics, such as the Area Under the Receiver Operating Characteristic curve (AUROC) and the Area Under Precision-Recall Curve (AUPRC) at two levels: observational level and encounter level. The observation level performance assessed the model's accuracy each time the patient's data was updated, whereas the encounter level performance evaluated the model's performance for the entire hospital encounter of a patient.

Results:
    Relatively promising results were obtained with the sliding window method and RNN with LSTM node.

Conclusions:
    The severe imbalance in the dataset posed a significant challenge for accurate prediction. Despite trying different methods, such as assigning different weights to positive and negative

observations on a binary cross-entropy loss function and utilizing a focal binary entropy loss that accounts for the imbalance structure, the performance did not significantly improve compared to the XGBoost model.

# Introduction

I. Background

The ICU provides critical care and life support to severely ill and injured patients in the hospital. Due to limited resources and high demand for ICU services, physicians must closely monitor patients in the ICU and intervene as soon as possible to prevent deterioration. To reduce ICU admission delays for deteriorating patients, physicians need to identify patients at high risk of deterioration. PICTURE is an analytical model that mimics this decision-making process and identifies patients at high risk of imminent ICU transfer, respiratory failure, or death using a traditional machine learning classification algorithm based on gradient-boosting decision trees (XGBoost).

The primary goal of this project was to investigate the potential of deep-learning models to replace XGBoost and achieve comparable or better predictive power. We aimed to explore the ability of deep-learning models to capture time and patient-wise dependencies that were not accounted for in the existing XGBoost model. Although PICTURE treated each observation independently, in reality, there are time-dependent trends that the existing model did not capture within one encounter.

From a technical perspective, we were interested in exploring different deep-learning models that allow us to capture time dependency within each encounter. We also aimed to decrease the false positive rate and improve the performance of two threshold-independent criteria (AUROC and AUPRC), which enable us to impartially evaluate the models without having to factor in the different hospitals' risk tolerance level reflected in the individual choice of thresholds.

II. Data Description

The PICTURE dataset was obtained from the Electronic Health Record (EHR) of adult patients (≥ 18 years) who were hospitalized in a general ward at the University of Michigan Hospital between 2014 and 2018. The dataset was collected from a large tertiary, academic medical system. To be eligible for inclusion in the study, patients had to be 18 years or older. The dataset was divided into training, validation, and test sets. We split both the windowed dataset and the granular dataset patient-wise into three subsets to ensure that each patient was only included in one set and maintain the independence of the training, validation, and test sets. The original data were in a long format, with each row corresponding to a new measurement at a new time point (Cummings et al, 2021). The dataset contains 46 independent variables, including vital signs (such as body temperature, pulse rate, respiration rate, and blood pressure), physiological observations (such as oxygen saturation), laboratory and metabolic values, and demographic information (such as gender and age). The dependent variable, or "target," was whether a patient experienced death, ICU transfer or accommodation, mechanical ventilation, or cardiac arrest within 24 hours. Appendix A provides summary statistics for features with large variance .

Because of the different frequency at which data was collected for each patient, the long format data were then simplified by grouping each encounter's data into 8-hour windows to ensure that each encounter would have the same number of data points for the same amount of time in the hospital. This approach helped prevent bias on encounters with more frequent updates while training the model. For clarity, we will refer to this aggregated data as the windowed dataset and the original long format data as the granular dataset.

In the windowed dataset's training set, there were 1,467,255 observations that included a total of 99,390 encounters. Out of these, there were 4,672 positive encounters, which signifies a deterioration in the patient's condition. The validation set of the same dataset contained 361,407 observations with 25,093 encounters, out of which 1,167 were associated with events. The test set included 1,521,108 medical records with 93,441 encounters, out of which 4,936 had events recorded.

There are 223, 824 encounters in the windowed dataset and 223, 640 in the granular dataset. Table 1 is the summary of the windowed dataset and granular dataset. Note that the number of encounters in different datasets were not matching between the windowed dataset and granular dataset. This is due to the fact that when we go to do the windowing, sometimes we lose patients who have been there a very short amount of time.

| Dataset | # Patients | Subset | # Medical Records | # Encounters | # Event | Event Rate |
|---|---|---|---|---|---|---|
| Windowed dataset | 132,589 | Training | 1,467,255 | 99,390 | 4,672 | 0.047 |
| | | Validation | 361,407 | 25,093 | 1,167 | 0.0465 |
| | | Test | 1,521,108 | 99,341 | 4,936 | 0.0497 |
| Granular Dataset | 132,474 | Training | 12,748,185 | 99,248 | 4,487 | 0.0452 |
| | | Validation | 3,183,698 | 25,051 | 1,112 | 0.0444 |
| | | Test | 1,521,108 | 93,441 | 4,936 | 0.0528 |

Table 1: Summary of dataset

We define the encounter length as the number of observations within this encounter. There was a large deviance of the encounter length in the dataset. Figure 1.1 shows the distribution of the encounter length in the windowed dataset while Figure 1.2 shows the distribution of the encounter length in granular dataset. The x-axis denotes the range of encounter length and y-axis denotes how many encounters there are in a particular range.

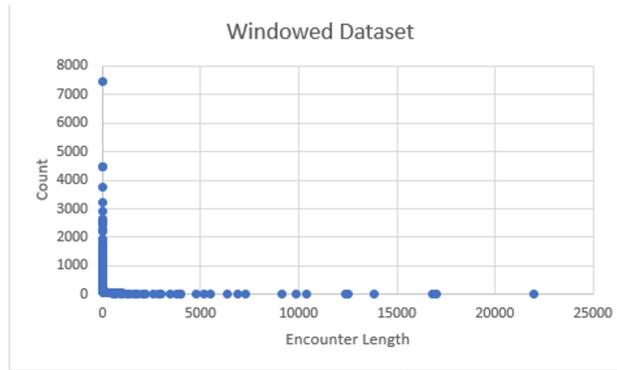
Figure 1.1 Windowed Dataset Encounter Description

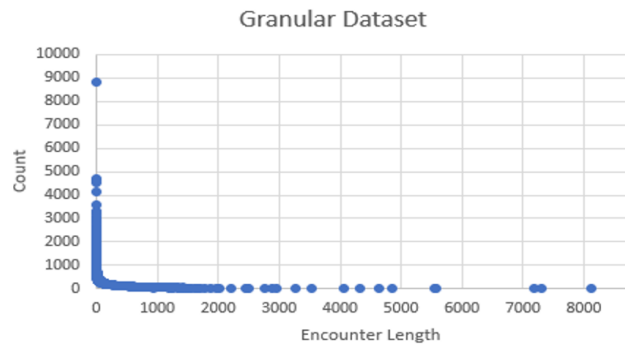
Figure 1.2 Granular Dataset Encounter Description

In terms of feature distributions, some of the features have large standard deviations, including age, systolic pressure, diastolic pressure, mean arterial pressure (MAP), pulse pressure, urine, weight and maximum oxygen supplementation within 24 hours. The detailed information about the features with large variance are provided in Appendix A.

# Related Works

Cummings et al. utilized the XGBoost model to validate the PICTURE model's ability to predict unexpected deterioration in general ward and COVID-19 patients (Cummings et al, 2021). The primary evaluation criteria for the models were AUROC and AUPRC, and the results were promising compared to the previous early warning system, Epic Deterioration Index (EDI).

Various data representations have been developed by different authors to enhance the models' performance. For example, Yu et al. introduced the bag-of-words representation for relevant medical events based on the most recent history as inputs. They also utilized latent semantic analysis (LSA) to encode patients' states into low-dimensional embeddings, employing various LSTM models' hidden states, self-attention mechanisms, and average pooling (Yu et.al 2020).

Ho et al. developed Learned Binary Masks (LBM) and KernelSHAP to generate an attribution matrix for identifying which EMR variables contributed to an RNN model's risk of mortality (ROM) predictions. They also proposed three aggregation methods, including aggregation over volatile time periods within individual patient predictions, over populations of ICU patients sharing specific diagnoses, and across the general population of critically ill children (Ho et al, 2021).

Several deep learning models were employed to capture the internal relationships across data. For instance, Yu et al. used RNN to learn input-output sequences by considering a hidden state, where bidirectional long short-term memory performed better by capturing both forward and backward temporal dependencies. Shukla and Marlin introduced multi-time attention networks for irregularly sampled time series, with a focus on the PhysioNet challenge dataset 2012, which comprises records from 12,000 ICU stays with about 42 extracted features, achieving an AUPRC of approximately 40%(Shukla et.al, 2021).

Cummings and Yu et al. employed AUPRC and AUROC to continuously monitor the patient's mortality risk and test the performance of the models.

# Methods

I. Data Preprocessing

To capture the time dependency of the data, we grouped all data points by encounter ID, which was defined by the "Contact Serial Number" (CSN). Afterward, we applied one-hot encoding to categorical data and standardized all continuous variables using standardization, which centers values around the mean and scales them to unit standard deviation. This approach helped to reduce the range difference of all variables, making it possible for the deep-learning model to train more efficiently.

The model implementation has only one flexible dimension, with three dimensions of input, including batch size, timestamps, and number of features. We can choose to keep the batch size or the timestamp fixed, depending on the situation. We experimented with three data preprocessing methods: sliding window, dense sliding window, and smart batching. The sliding window and dense sliding window methods aimed to keep the timestamp fixed, while smart batching aimed to keep the batch size fixed.

To capture time dependency between health records, we experimented with the sliding window method as our first data preprocessing approach. The aim of this method was to keep the timestamp fixed without padding too many zeros. Compared with padding the timestamps of all inputs with the maximum number of records within one encounter, this approach helped save memory and reduce sparsity. Our input was an array containing all features of one encounter from the windowed dataset. To use this method, we selected a hyperparameter called timestamp, which defined the time window that we were interested in. Each data point in the windowed dataset was collected every 8 hours, so the time window was set to (#Timestamp * 8) hours. Next, we collected the features within this time window to make predictions. Since the length of encounters varied greatly across different patients, we encountered two situations as shown in Figure 2 and Figure 3. Encounters with fewer medical records than timestamps were defined as short encounters, while those with more records were defined as long encounters. Here T denoted the time points.

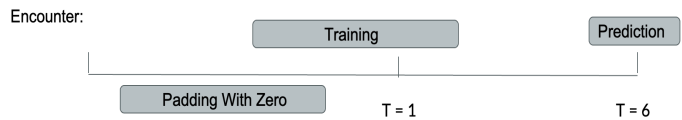

Figure 2: Sliding Window Method Illustration for Short Encounters

In terms of short encounters, we would pad with zero at the beginning. Since we would mask all zeros during training, this would not have a negative impact on the prediction. The feature would be in the shape of [1, 6, 46]. The "target" of the last data points, which is T=6 on Figure 2, would be the label for this prediction.

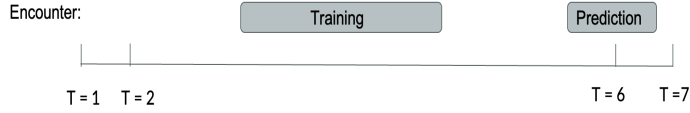

Figure 3: Sliding Window Method Illustration for Long Encounters

In terms of long encounters, suppose there were N medical records within this encounter, we would first form arrays in the form of $[X_i, X_{i+1}, X_{i+2}, X_{i+3}, X_{i+4}, X_{i+5}]^T)$ ($i = 1 \cdots N - timestamp + 1$), where $X_i$ is the selected feature of a medical record in this encounter. The array would be in the shape of [Timestamp, number of features]. Then, we would stack N-timestamp+1 arrays together, resulting in the dimension to be [N-timestamp+1, timestamp, number of features], to form an input array. Therefore, on Figure 3, the feature would be in the shape of [2, 6, number of features].

The second method was the dense sliding window method. Similarly, we would also pick a hyperparameter, timestamp to define the time window. Compared with the sliding window method, this method would also pad some zeros for long encounter cases and merge long encounter cases and short encounter cases together. This made more sense as in real life, we keep getting new medical records and would make predictions for the next time point. We would not know whether an encounter would be a short encounter or a long encounter beforehand. If there were N medical records, we would get N arrays in the form of $[X_1, X_2, \cdots X_i]^T$ where i = 1 $\cdots$ N and $X_i$ was the selected feature of a medical record in this encounter. We would pad zeros at the beginning of the array to make sure that the number of medical records within one array would equal the size of the timestamp. After that, we would stack those N arrays together to form an input array with the size of. Figure 4 would be an example of the dense sliding window method.

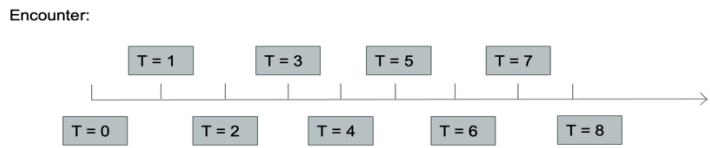

Figure 4: Dense sliding window method

Here T denoted time points. Suppose there were 8 data points within this encounter, we would form 8 arrays where $[X_1, X_2 \cdots X_i]$, i = 1,2, $\cdots$ 8 and $X_i$ is a medical record in this encounter. The labels were defined as the target of the last medical records. Finally, we stacked them together to form an input array with the size of [8, 6, number of features].

The third method utilized is the smart batching method, which was adapted from an NLP taskm(Chris, 2020). This method aimed to maintain a fixed batch size while keeping the size of the timestamp fixed. The advantage of this method is that it eliminates the need to know the size of the time window beforehand. The dataset used for this method was the granular dataset, where each row corresponded to a new measurement at a new time point. To track the feature with respect to time, a time difference column was added and passed as another feature to the model. Uniform length batching was used to minimize the amount of padding with zeros. An example of how padding was done in this case can be seen in Table 2.

|  | 1 | 2 | 3 | 4 | 5 | 6 | 7 | 8 | 9 | 10 | 11 | 12 | 13 | 14 |  |
|---|---|---|---|---|---|---|---|---|---|---|---|---|---|---|---|
| Encounter1 | DP1 | DP2 | DP3 | DP4 | DP5 | DP6 | PAD | PAD |  |  |  |  |  |  | Batch Length = 8 |
| Encounter2 | DP1 | DP2 | DP3 | DP4 | DP5 | DP6 | DP7 | PAD |  |  |  |  |  |  |  |
| Encounter3 | DP1 | DP2 | DP3 | DP4 | DP5 | DP6 | DP7 | DP8 |  |  |  |  |  |  |  |
| Encounter4 | DP1 | DP2 | DP3 | DP4 | DP5 | DP6 | DP7 | DP8 |  |  |  |  |  |  |  |
| Encounter5 | DP1 | DP2 | DP3 | DP4 | DP5 | DP6 | DP7 | DP8 | PAD | PAD |  |  |  |  | Batch Length = 10 |
| Encounter6 | DP1 | DP2 | DP3 | DP4 | DP5 | DP6 | DP7 | DP8 | PAD | PAD |  |  |  |  |  |
| Encounter7 | DP1 | DP2 | DP3 | DP4 | DP5 | DP6 | DP7 | DP8 | PAD | PAD |  |  |  |  |  |
| Encounter8 | DP1 | DP2 | DP3 | DP4 | DP5 | DP6 | DP7 | DP8 | DP9 | DP10 |  |  |  |  |  |
| Encounter9 | DP1 | DP2 | DP3 | DP4 | DP5 | DP6 | DP7 | DP8 | DP9 | DP10 | DP11 | DP12 | DP13 | PAD | Batch Length = 14 |
| Encounter10 | DP1 | DP2 | DP3 | DP4 | DP5 | DP6 | DP7 | DP8 | DP9 | DP10 | DP11 | DP12 | DP13 | PAD |  |
| Encounter11 | DP1 | DP2 | DP3 | DP4 | DP5 | DP6 | DP7 | DP8 | DP9 | DP10 | DP11 | DP12 | DP13 | DP14 |  |
| Encounter12 | DP1 | DP2 | DP3 | DP4 | DP5 | DP6 | DP7 | DP8 | DP9 | DP10 | DP11 | DP12 | DP13 | DP14 |  |

Table 2: Smart Batching Illustration

To preprocess the data, we first grouped all the medical records belonging to one encounter and formed an encounter list. Then, we sorted the encounter list by encounter length to ensure that adjacent encounters had similar lengths. Next, we randomly selected a fixed number of encounters and formed a batch of input. This approach reduced the number of zeros we needed to pad, as the number of padded zeros depended on the maximum length within that batch. Each encounter would have a label. Each batched input would have # batch size labels in total.

II. Deep Learning Models:
- Recurrent Neural Network:

A recurrent neural network (RNN) is any network that contains a cycle within its network connections, meaning that the value of some unit is directly, or indirectly, dependent on its own earlier outputs as an input (Jurafsky and martin, 2009). It is hard for vanilla RNN to make use of information distant from the current point of processing. Long short-term memory(LSTM) Network is an extension to RNN to manage the task of maintaining relevant context over time. The common structure of RNN is shown on Figure 5. $X^i$ denotes i-th input, $W_h$ denotes the weight. $\hat{y}$ denotes the predicted label, $J^i(\theta)$ denotes the loss.

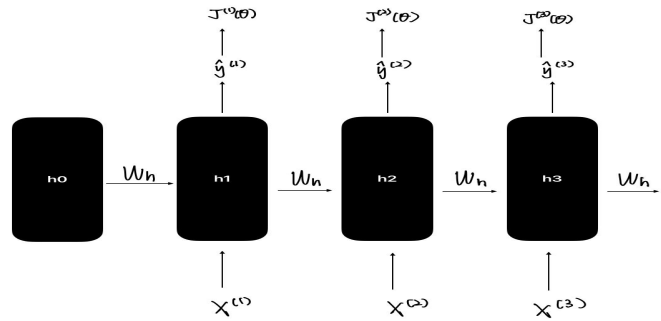

Figure 5 RNN Structure

At every time step, we can unfold the network for k time steps to get the output at time step k+1. $x_t$ denotes the t-th input. $h_t$ denotes t-th hidden state. $w_x$ denotes the weight for input. $w_h$ denotes the weight for the hidden state. f is an activation function. We can represent each hidden state:

$$h_{t+1} = f(x_t, h_t, w_x, w_h, b_h) = f(w_t x_t + w_h h_t + b_h)$$

The output y at time t is computed as:

$$y_t = f(h_t, w_y) = f(w_y h_t + b_y)$$

- Transformer:

The Transformer model has demonstrated remarkable performance in various natural language processing and computer vision tasks. Its ability to capture long-range dependencies and interactions makes it particularly appealing for time series modeling. In comparison to Recurrent Neural Networks (RNNs), Transformers offer several advantages.

Firstly, while RNNs process input sequences sequentially, Transformers can process all elements of a sequence in parallel, leading to more efficient training and inference times. Secondly, RNNs may struggle with capturing long-range dependencies in a sequence, as information from earlier time steps can get diluted or lost over time. Transformers address this issue through self-attention mechanisms that enable any element in the sequence to attend to any other element, more effectively capturing long-range dependencies. Finally, RNNs can suffer from the vanishing gradient problem, where gradients can become very small during backpropagation, hindering the network's ability to learn long-term dependencies. Transformers do not face this issue, as they do not rely on sequential processing.

The transformer architecture consists of an encoder and a decoder, both of which are composed of multiple layers of self-attention and feedforward neural networks. The input sequence is fed into the encoder, which consists of N identical layers. Each layer has two sublayers:1) Self-attention: The input sequence is transformed into three sequences - query, key and value vectors, which are then used to compute a weighted sum of the values, where the weights are determined by the similarity between the query and key vectors. 2) Feedforward:

The output of the self-attention layer is fed through a fully connected feedforward network. The decoder is also composed of N identical layers, each with three sub-layers: 1) Masked self-attention: Similar to self-attention in the encoder, but only considers the previous tokens in the sequence.2) Encoder-decoder attention: The query vectors from the decoder are matched with the key and value vectors from the encoder, allowing the decoder to attend to different parts of the input sequence. The output of the final decoder layer is fed through a linear layer and a softmax/sigmoid activation function to produce the final output sequence.

III.     Training scheme:

To ensure that we did not touch the real test dataset until we had a relatively good model, we divided our original training dataset, consisting of data from 2014 to 2018, into three subsets: training-training, training-validation, and training-test. The training-training subset contained 60% of the data, the training-validation subset contained 20%, and the remainder was placed in the training-test set. Once we had a reasonably good model, we also tested it on the original validation dataset and the original test dataset.

We observed a significant difference between positive and negative samples, so we employed class weights when training the model. Class weights assigned a weight to each class that was inversely proportional to the class frequency, meaning that the minority class had a higher weight than the majority class. During training, the loss function was computed using these weights, which gave more importance to the minority class samples and increased the likelihood of correctly classifying them. We calculated the class weights as follows:

$$weight\ of\ zero\ class\ = \frac{\#\ total\ labels\ in\ the\ dataset}{\#\ zero\ labels\ in\ the\ dataset\ *\ 2}$$

$$weight\ of\ one\ class\ = \frac{\#\ total\ labels\ in\ the\ dataset}{\#\ one\ labels\ in\ the\ dataset\ *\ 2}$$

There were two loss functions used for the experiment. Binary Cross-Entropy is commonly used for binary classification. Binary cross-entropy is a simple and fast loss function to compute during training. It penalizes the model more heavily for confident mistakes than for uncertain ones. This means that the model is incentivized to be more cautious in its predictions, which can lead to better generalization performance.

$$H_p(q) = -\frac{1}{N}\sum_{i=1}^{N} y_i \cdot \log(p(y_i)) + (1 - y_i) \cdot \log(1 - p(y_i))$$

Note: N: number of samples, $y_i$: predicted value, p: probability

However, binary cross-entropy suffered from class imbalance, which was a major issue in this project. Therefore, we also implemented focal loss to solve this problem. Focal loss assigned higher weights to misclassified examples of the minority class, which can help to balance the influence of the two classes. In addition, focal loss placed more emphasis on hard examples, which were examples that were difficult to classify correctly. By doing so, it helped the model to focus more on the examples that are most important for improving performance. The focal loss function modified the binary cross-entropy loss function by introducing a modulating factor that down-weights the contribution of easy examples and up-weights the contribution of hard examples. The focal binary entropy was defined as follows:

$$FL(py) = -(1-p)^\gamma \cdot y \cdot log(p) - p^\gamma \cdot (1-y) \cdot log(1-p)$$

where:

$\gamma$ was a focusing parameter that controlled the degree of down-weighting for easy examples.

$(1-p)^\gamma$ and $p^\gamma$ were the modulating factors that adjust the contribution of each example based on its difficulty. Hard examples (i.e., examples with low predicted probabilities) were up-weighted, while easy examples (i.e., examples with high predicted probabilities) were down-weighted. $\hat{y}$ was the predicted value. p denoted probability function.

Two different optimizers were used in the study. Root Mean Square Propagation (RMSProp) was employed as a variant of stochastic gradient descent (SGD), which adapted the learning rate for each weight based on the average of the squared gradients. It could converge faster and avoid oscillations compared to regular SGD, especially on problems with sparse gradients. However, RMSProp was sensitive to the choice of hyperparameters and might converge to a suboptimal solution when the learning rate was too high or too low. In addition, Adam was also used as a variant of stochastic gradient descent (SGD), which combined ideas from RMSProp and momentum to achieve faster convergence and better generalization performance.

In comparison to RMSProp, Adam's adaptive learning rate also considered the first and second moments of the gradients, resulting in more accurate and efficient updates. However, Adam was less sensitive to the choice of hyperparameters than RMSProp and was less likely to converge to a suboptimal solution.

To avoid overfitting and improve generalization, we utilized a dropout layer, which is a regularization technique commonly used in neural networks. The dropout layer randomly drops out some of the neurons in the network during training, encouraging the network to learn more generalized features that are less dependent on specific neurons. We also implemented early stopping to prevent overfitting. This involved monitoring the model's performance on a validation set and stopping the training process when the performance began to deteriorate.

- Model 1: LSTM Model
    - Experiment 1.1: Preprocessing with sliding window method

After the windowed dataset had been processed using a sliding window method, a model structure was built as depicted in Figure 6 for training the model. The deep learning structure used in this model, also shown in Figure 6, began with a mask layer to mask all padded zeros. Then several blocks containing one LSTM layer, one Layer Normalization layer, and one dropout layer were appended. Finally, a dense layer was added at the end. Masking is often used in models that process variable-length input sequences. The mask layer was used to indicate which parts of the input sequence are valid and which parts should be ignored. Layer normalization was employed to address internal covariate shift, which is a change in the distribution of the input values to a layer during training. It also helped to stabilize activations and improve generalization. LSTM was utilized to learn time dependency features. The final dense layer mapped the learned features to a set of class probabilities.

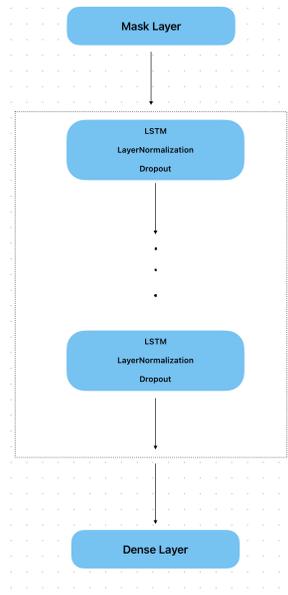

Figure 6 Model Structure 1

We experimented with both Adam and RMSProp optimizers and combined with two loss functions, binary cross-entropy and focal binary cross-entropy. When using focal binary cross-entropy, we tried several gamma values to determine the one with the best performance. Additionally, we weighted the loss using weights calculated from the training dataset of the windowed dataset to train the model. We set the weight for predictions being 1 to 62.71 and the weight for predictions being 0 to 0.50.

- ❖ Experiment 1.2: Preprocessing with dense sliding window method + LSTM model

After processing the windowed data with a dense sliding window method, we trained a model with the same class weights and model structure as shown in Figure 6. We experimented with both Adam and RMSProp optimizers, using two loss functions - binary cross-entropy and focal binary cross-entropy - for comparison. The weights used were the same as those in Experiment 1.1.

❖ Experiment 1.3: Preprocessing with smart batching method + LSTM model

After the granular dataset was processed using a smart batching method, the model was trained with the structure depicted in Figure 6. The loss function was weighted with the weights calculated from the training dataset of the granular dataset to train the model. The weight for predictions being 1 was set to 42.67, while the weight for predictions being 0 was set to 0.51.

- Model 2: Transformer Method:

The transformer was built by starting with two encoders that were connected with one dense layer and one dropout layer. The multi-head attention mechanism allowed the model to attend to different parts of the input sequence with different sets of learned weights, enabling it to capture different relationships between the input and output. Layer normalization was employed to stabilize activations and improve generalization. The Encoder Structure was built according to Figure 7.

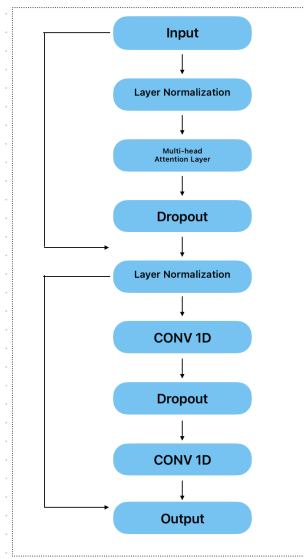

Figure 7 Transformer Encoder Structure

❖ Experiment 2.1: Preprocessing with dense sliding window method

After the windowed dataset was processed using a dense sliding window method, we built a Transformer to train the model. Both Adam and RMSProp optimizers were tried. Two loss functions, binary cross-entropy and focal binary cross-entropy, were used for comparison. The same weights as those in experiment 1.1 were used.

❖ Experiment 2.2: Preprocessing with smart batching method

After processing the granular dataset with a smart batching method, a Transformer was trained. The loss was weighted, and the weights were calculated from the training dataset of the granular dataset to train the model. The weight for predictions being 1 was set to 42.67, and the weight for predictions being 0 was set to 0.51.

IV. Evaluation Method

The performance was assessed on two scales: observation level and encounter level. We mainly focused on AUPRC and AUROC.

The AUPRC represents the average precision across the range of possible sensitivities, with precision being defined as the probability that a patient deteriorates when the model sends an alert, and sensitivity being the proportion of deteriorations that were caught by the model.

The AUROC score similarly provides a threshold-independent measure of performance, this time in terms of sensitivity and specificity. Notably, an AUROC score of 1 is a perfect score and an AUROC score of 0.5 corresponds to random guessing.

# Results

I. Baseline:

The baseline we pick is the XGBoost model described in Cummings, 2021. Table 3 summarizes the performance of this model. Since using an RNN has considerable advantages over XGBoost to capture time dependence, the goal of this project is to at least make a comparable model to XGBoost.

| Model | Original Dataset | Level | AUROC | AUPRC | Event Rate |
|---|---|---|---|---|---|
| XGBoost | Training | Observation | 0.8609 | 0.179 | 0.00797 |
| | | Encounter | 0.8476 | 0.44 | 0.047 |
| | Validation | Observation | 0.825 | 0.125 | 0.008 |
| | | Encounter | 0.827 | 0.387 | 0.0465 |
| | Test | Observation | 0.84 | 0.133 | 0.00826 |
| | | Encounter | 0.8321 | 0.3723 | 0.0528 |

Table 3 XGBoost Result

II. Deep Learning Model:
- ❖ Experiment 1.1:

We tried different numbers of LSTM→LayerNormalization→Dropout block and different optimizers and loss functions. We only trained on a subset of the original training dataset to avoid overfitting. In addition, we set patience of early stopping to 10 and the patience of reducing learning rate of reducePlateau was 6 to avoid overfitting. By stopping the training when the validation loss stops improving, we can prevent the model from overfitting. The model that gave us the highest AUROC and AUPRC has the structure on Figure 8.

The time window we set was a week. The raw data was grouped into 8 hour windows for the windowed dataset, meaning that there are 3 windows in a day, and therefore 21 in a week. If we used slide window method, the structure of the best model we found is:

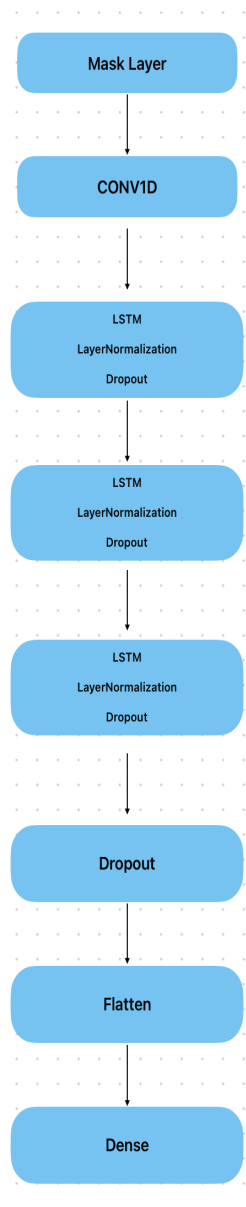

Figure 8 Experiment 1.1 Structure

We trained on the training-training subset and validated on the training-validation subset. Next, we tested the model on both the training-test dataset and a real validation dataset. The results showed a significant improvement over the XGBoost model on the training-training and training-validation datasets. However, when we tested the model on the original validation and test sets, it did not perform well, indicating the problem of overfitting. According to Table 4, the event rate of the training dataset was much higher than that of other subsets, which suggests that the overfitting occurred because the training data was not representative of the underlying pattern in the data. In this case, the model overemphasized certain features or patterns in the training data, leading to poor generalization performance on new and unseen data.

| Model | Data Preprocessing | Original Dataset | Dataset | Level | AUROC | AUPRC | Event Rate |
|---|---|---|---|---|---|---|---|
| RNN | Fixed Slide Window | Training | Training | Observation | 0.905 | 0.355 | 0.1023 |
| | | | | Encounter | 0.942 | 0.47 | 0.0395 |
| | | | Validation | Observation | 0.872 | 0.312 | 0.01 |
| | | | | Encounter | 0.894 | 0.432 | 0.0393 |
| | | Validation | Validation | Observation | 0.75 | 0.054 | 0.00816 |
| | | | | Encounter | 0.78 | 0.208 | 0.04652 |
| | | Test | Test | Observation | 0.709 | 0.055 | 0.00826 |
| | | | | Encounter | 0.585 | 0.088 | 0.0528 |

Table 4 : RNN + Slide Window Method Result

In order to find the reason why the training data is not representative of the underlying pattern in the data, we also split the dataset into two parts: long encounters and short encounters and trained separately. We found that the performance is better for short encounters. That is to say, there might be some flaws on padding for long encounters. Therefore, we switched to a dense sliding window method.

❖ Experiment 1.2:

If the dense sliding window method was used, the structure of the best model that was found had a masking layer and 5 LSTM layers with batch normalization in between. The model was trained on the original training dataset by splitting it into a training-training subset and a training-validation subset. However, the results did not outperform XGBoost. As a result, we tested the model only on the original validation dataset. The results are presented in Table 5.

| Model | Data Preprocessing | Original Dataset | Dataset | Level | AUROC | AUPRC | Event Rate |
|---|---|---|---|---|---|---|---|
| RNN | Dense Sliding Window | Training | Training | Observation | 0.749 | 0.036 | 0.00795 |
| | | | | Encounter | 0.794 | 0.175 | 0.0468 |
| | | | Validation | Observation | 0.715 | 0.021 | 0.0077 |
| | | | | Encounter | 0.732 | 0.134 | 0.0461 |
| | | Validation | Validation | Observation | 0.735 | 0.021 | 0.00816 |
| | | | | Encounter | 0.693 | 0.136 | 0.0465 |

Table 5: RNN + Dense Sliding Window Result

We can see that overfitting didn't happen in this case. However, the AUROC and AUPRC were relatively low compared to the baseline. One reason might be that there was insufficient training data in this model. We only trained on 60% of the training data in XGBoost model training. The second reason might be that the data preprocessing method might need to be improved. The third reason can be that the imbalance might still not be solved even with class weights in training.

❖ Experiment 1.3:

The AUROC is about 55% and AUPRC is about 4%. One of the reasons for the worse performance compared with the other two methods is that we are using granular dataset for this method. The windowed dataset that was used in the previous model has been simplified by grouping each encounter's data into 8-hour windows to ensure that each encounter would have the same number of data points for the same amount of time in the hospital. The granular dataset might suffer from the bias on encounters with more frequent updates while training the model. Thus, the performance would be worse than the previous two methods.

❖ Experiment 2.1:

We tried several hyperparameter combinations to find the best model, including setting the number of encoders to 1, 2, or 3, the size of the transformer head to 128, 256, or 512, and the number of transformer heads to 4, 6, or 8. After completing the hyperparameter search, we discovered that the best performance on AUROC and AUPRC was achieved when the transformer head was set to 128, the number of transformer heads was set to 6, the feedforward dimensions were set to 64, and the model included only 2 transformer blocks (as shown in Figure 9). The transformer encoder was defined as what Figure 7 showed.

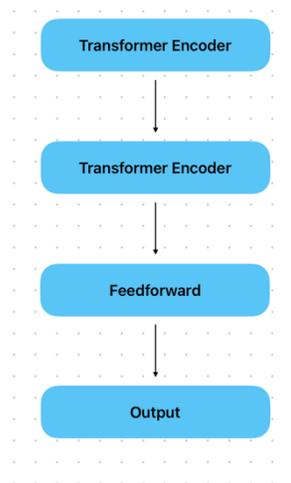

Figure 9: Experiment 2.1 Structure

We trained the model on the training-training subset and validated it on the training-validation subset. However, the results did not outperform XGBoost, so we only tested the model on the original validation dataset. The results are shown in Table 6.

| Model | Data Preprocessing | Original Dataset | Dataset | Level | AUROC | AUPRC | Event Rate |
|---|---|---|---|---|---|---|---|
| Transformer | Dense Sliding Window | Training | Training | Observation | 0.749 | 0.023 | 0.008102 |
| | | | | Encounter | 0.725 | 0.127 | 0.0487 |
| | | | Validation | Observation | 0.688 | 0.021 | 0.00678 |
| | | | | Encounter | 0.711 | 0.114 | 0.04204 |
| | | Validation | Validation | Observation | 0.76 | 0.067 | 0.0092 |
| | | | | Encounter | 0.761 | 0.241 | 0.05 |

Table 6  Transformer + Dense Sliding Window Result

The AUROC and AUPRC were relatively low compared to the baseline. There are a couple of reasons for this. The Transformer model is a relatively complex model. We probably need more features to help the model to learn. In addition, the data preprocessing method might need to be improved.

❖ Experiment 2.2:

The AUROC is about 60% and AUPRC is about 9%. Here we were using the granular dataset to train the model. The granular dataset might suffer from the bias on encounters with more frequent updates while training the model.

# Conclusion and Future Directions

Previous research on the PICTURE dataset was conducted using the XGBoost model. This project aimed to benchmark the XGBoost model against deep learning models. To match the implementation of the deep learning models, we utilized three different methods for data processing, including the slide window method, dense sliding window method, and smart batching method. The slide window and dense sliding window methods were applied to the windowed dataset, while the smart batching method was used on the granular dataset. Each method had its pros and cons, with the slide window method producing relatively good results but requiring more prior knowledge for correct predictions. Recurrent Neural Network with LSTM nodes and Transformer models were developed using these preprocessing methods; however, their performance was inferior to the baseline model. During the exploration phase, we noticed that the deep learning models predicted "deterioration" at the beginning but later predicted "no deterioration," which conflicted with the label's actual meaning. Additionally, the imbalanced dataset posed challenges for accurate predictions, even with class weights.

Smart batching is a new data processing method that is closer to real-time conditions. We found that simpler model structures worked better for smart batching. The next step is to improve smart batching methods by decreasing model complexity and adjusting parameters. Additionally, new model structures, such as multi-time attention networks for irregularly sampled time series, can be implemented to enhance performance. Furthermore, we suggest two new methods to address the problem of imbalance. The first is ensemble learning, in which multiple models are trained on different subsets of the imbalanced dataset and their predictions combined to obtain a final prediction. This approach can improve the model's generalization performance. The second method is transfer learning, which involves using a pre-trained model on a large dataset as a feature extractor and fine-tuning it on the imbalanced dataset. This can help the model learn better data representations.

# References

1. Yu K, Zhang M, Cui T, Hauskrecht M. Monitoring ICU Mortality Risk with A Long Short-Term Memory Recurrent Neural Network. Pac Symp Biocomput. 2020;25:103-114. PMID: 31797590; PMCID: PMC6934094.
2. Long V. Ho, Melissa Aczon, David Ledbetter, and Randall Wetzel. 2021. Interpreting a recurrent neural network's predictions of ICU mortality risk. *J. of Biomedical Informatics 114, C* (Feb 2021).
3. Cummings BC, Ansari S, Motyka JR, Wang G, Medlin RP et.al. Predicting Intensive Care Transfers and Other Unforeseen Events: Analytic Model Validation Study and Comparison to Existing Methods. *JMIR Med Inform. 2021 Apr 21;* 9(4):e25066. doi:10.2196/25066. PMID: 33818393; PMCID: PMC8061893.
4. Shukla, S. N., & Marlin, B. M. Multi-time attention networks for irregularly sampled time series. *International Conference on Learning Representations (ICLR), 2021*
5. Smart Batching Tutorial – Speed up BERT training, Chris McCormick, 2020, http://mccormickml.com/2020/07/29/smart-batching-tutorial/.
6. Dan Jurafsky and James H. Martin, Speech and Language Processing, 2nd Edition, Prentice Hall, 2009.

# Appendix A

|  | Granular Test |  | Granular Valid |  | Granular Train |  |
|---|---|---|---|---|---|---|
| Feature Name | Median | IQR | Median | IQR | Median | IQR |
| Age | 61 | [47, 71] | 61 | [48, 70] | 61 | [49, 70] |
| Diastolic Pressure | 67 | [59, 76] | 67 | [59, 75] | 67 | [59, 75] |
| Mean Arterial Pressure | 87 | [77.66666667, 97] | 86.33333333 | [77, 96.33333333] | 86.66666667 | [77.33, 96.67] |
| Pulse Pressure | 57 | [47, 70] | 57 | [46, 69] | 57 | [47,70] |
| Urine | 278.9997014 | [150, 400] | 286.0987381 | [150, 400] | 289.6038631 | [150,400] |
| Weight | 177.91125 | [146.605625, 214.286875 ] | 179 | [147.2675, 214] | 178.793125 | [148.810625, 212.2] |
| Max oxygen supplementation within 24 hours | 1 | [0,2] | 0 | [0,2] | 0 | [0,2] |
| Systolic Pressure | 125 | [111., 140.] | 125 | [110., 140.] | 125 | [111., 141. ] |

|  | Windowed Test |  | Windowed Valid |  | Windowed Train |  |
|---|---|---|---|---|---|---|
| Feature Name | Median | IQR | Median | IQR | Median | IQR |
| Age | 61 | [47, 71] | 61 | [47, 71] | 61 | [48, 70] |
| Diastolic Pressure | 67 | [59, 76] | 67 | [59, 76] | 67 | [59, 76] |
| Mean Arterial Pressure | 87 | [77.67, 97] | 86.33 | [77.33, 96.33] | 86.67 | [77.67, 96.67] |
| Pulse Pressure | 57 | [47,70] | 57 | [47,68] | 57 | [47,69] |
| Urine | 278.9997014 | [150, 400] | 291.0171246 | [150, 400] | 300 | [150, 400] |
| Weight | 177.91125 | [146.605625,214.286875] | 176.368125 | [145.283125,211.5] | 176.7612259 | [146.165, 210.49923967] |
| Max oxygen supplementation within 24 hours | 1 | [0,2] | 1 | [0,2] | 1 | [0,2] |
| Systolic Pressure | 125 | [111., 141. ] | 124 | [110., 140.] | 125 | [111, 140] |

Note: IQR: [75 percentile value, 25 percentile value]